\documentclass[a4paper,fleqn]{cas-dc}

\usepackage[numbers]{natbib}

\usepackage{hyperref}
\usepackage{kotex}
\usepackage{xcolor}
\newcommand{\kb}[1]{\textcolor{black}{#1}}

\newcommand{\ie}{\textit{i}.\textit{e}.}
\newcommand{\eg}{\textit{e}.\textit{g}.}
\newcommand{\Fref}[1]{Figure \ref{#1}}

\newcommand{\Tref}[1]{Table \ref{#1}}

\newcommand{\Eref}[1]{Eq. \ref{#1}}

\newcommand{\degree}{\ensuremath{^\circ}}

\def\tsc#1{\csdef{#1}{\textsc{\lowercase{#1}}\xspace}}
\tsc{WGM}
\tsc{QE}
\tsc{EP}
\tsc{PMS}
\tsc{BEC}
\tsc{DE}

\begin{document}
\let\WriteBookmarks\relax
\def\floatpagepagefraction{1}
\def\textpagefraction{.001}
\shorttitle{ArrowGAN : Learning to Generate Videos by Learning Arrow of Time}
\shortauthors{Kibeom Hong et~al.}

\title [mode = title]{ArrowGAN : Learning to Generate Videos by Learning Arrow of Time}                      


\author[1]{Kibeom Hong}[type=editor,
                        orcid=0000-0003-2366-6318]
\cormark[1]

\credit{Writing - original draft, Conceptualization, Methodology, Software}

\address[1]{Department of Computer Science, Yonsei University, Seoul, Korea}

\author[2]{Youngjung Uh}[type=editor,
                        ]
\cormark[1]

\credit{Writing - original draft, Conceptualization, Methodology, Software}

\address[2]{Clova AI Research, NAVER, Korea}

\author[1]{Hyeran Byun}[type=editor,
                        ]
\cormark[2]
\credit{Supervision, Writing - review \& editing}

\cortext[cor1]{indicates equal contribution}
\cortext[cor2]{indicates corresponding author}


\begin{abstract}
Training GANs on videos is even more sophisticated than on images because videos have a distinguished dimension: time. While recent methods designed a dedicated architecture considering time, generated videos are still far from indistinguishable from real videos. In this paper, we introduce ArrowGAN framework, where the discriminators learns to classify arrow of time as an auxiliary task and the generators tries to synthesize forward-running videos. We argue that the auxiliary task should be carefully chosen regarding the target domain. In addition, we explore categorical ArrowGAN with recent techniques in conditional image generation upon ArrowGAN framework, achieving the state-of-the-art performance on categorical video generation. Our extensive experiments validate the effectiveness of arrow of time as a self-supervisory task, and demonstrate that all our components of categorical ArrowGAN lead to the improvement regarding video inception score and Fr\'echet video distance on three datasets: Weizmann, UCFsports, and UCF-101.
\end{abstract}


\maketitle

\begin{figure*}
\centering
\includegraphics[width=1.0\textwidth,keepaspectratio]{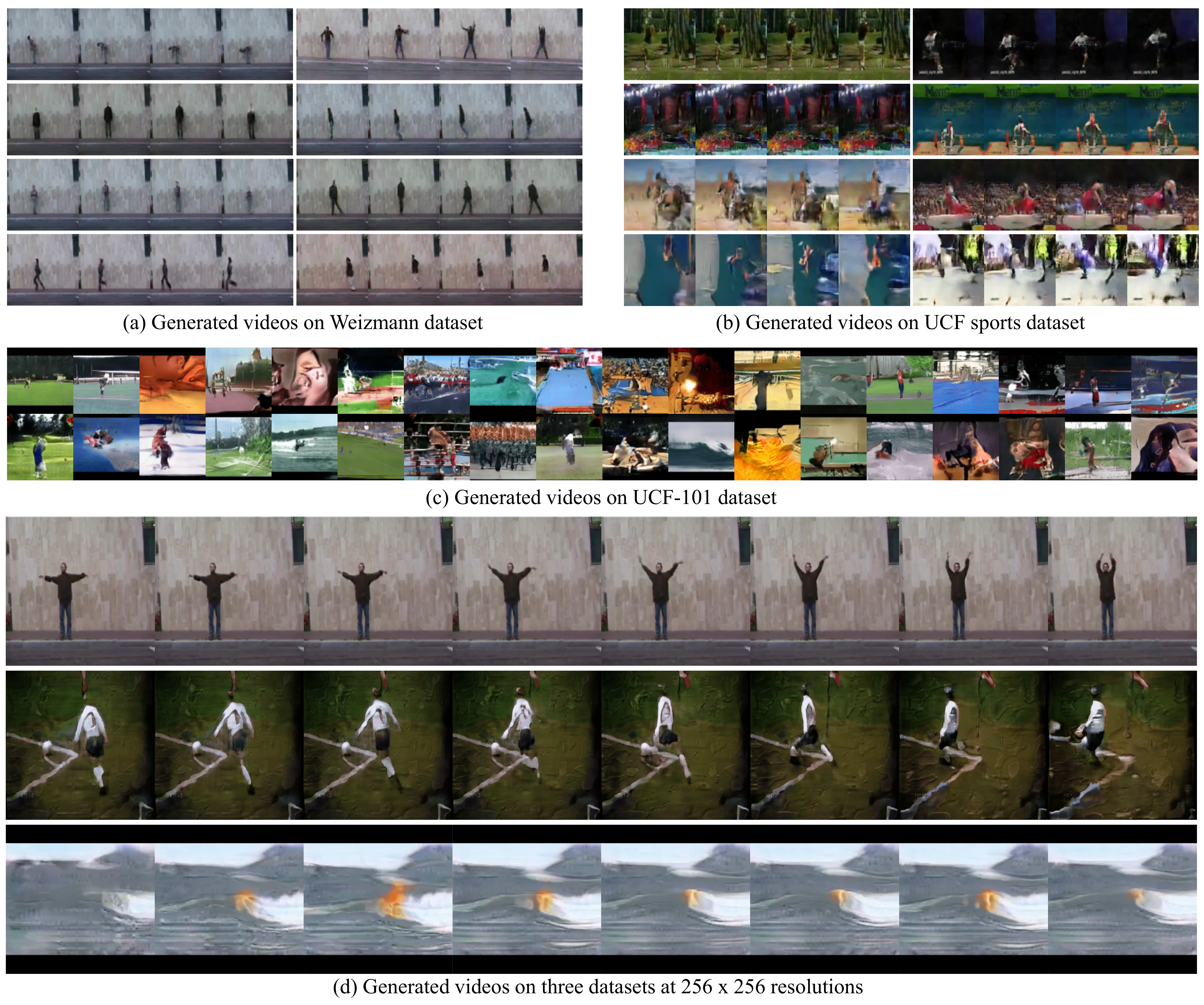}
\caption{Uncurated set of videos produced by our categorical ArrowGAN. (a,b) shows four consecutive frames for different 8 classes. (c) shows randomly sampled frames from each class in UCF-101 dataset. Also (d) illustrates eight sequential frames from generated videos of three datasets at 256x256 resolutions. (Best viewed in color.)}
\label{fig:teaser}
\end{figure*}

\section{Introduction}
Although recent methods in image generation~\cite{karras2019style,brock2018large} synthesize real-like images with or without hints, generating a video is more challenging than synthesizing a single image due to the complexity of videos. Several works have tried to generate videos with dense hints: predicting future frames from a starting frame~\cite{babaeizadeh2018stochastic,denton2018stochastic} and translating a sequence of segmentation maps into realistic videos~\cite{Wang2018VideotoVideoS,pan2019video}. On the other hand, GAN-based methods~\cite{vondrick2016generating,saito2017temporal,tulyakov2018mocogan} try to adopt success from image generation methods by handling temporal information with three-dimensional convolutions or sequential models~\cite{chung2014empirical} on the unconditional environment. However, they still have difficulty in generating recognizable motions.

In this paper, we focus on \textit{time}, which is the biggest difference between videos and images, and is the source of the increased data complexity. It overwhelms the power of discriminator classifying real and fake samples, which is the basis of GAN-based methods. Hence, we regard time as the key to solve the problem, and propose arrow-of-time discriminator (Arrow-D) which equips the sense of time.
  
Typically, humans have a common sense of temporal direction of motions, thus they can tell whether a given video is running forward or backward. For example, people run forward, balls fly into goals and water flows downstream, but not the reverse way. Likewise, neural networks also can recognize \textit{Arrow of Time} (AoT), whether a video is running forwards or backwards~\cite{wei2018learning}.

To this end, we demand Arrow-D to classify AoT as an auxiliary task. It gives extra advice for the generators to produce more plausible videos, based on the inductive bias that motions in generated videos should run forward in time and the learned spatio-temporal cues that activate forward output of the Arrow-D. We name the GAN frameworks with Arrow-D as \kb{ArrowGAN}.

Although previous studies~\cite{odena2017conditional,miyato2018cgans} showed that imposing auxiliary tasks to the discriminators \kb{improve} the quality of generated samples, they require human supervision such as class labels. On the other hand, Arrow-D can learn AoT in self-supervised manner because backward-running samples can easily be obtained by reversing training samples. Also, predicting AoT goes along with the original GAN objective in that it also is a binary classification between real (normal) and fake (reversed) samples.

We also tried predicting rotation~\cite{chen2019self,gidaris2018unsupervised} and predicting whether frames are shuffled or not~\cite{Misra2016ShuffleAL} which are shown useful in various tasks. While learning AoT is beneficial to generating videos, they do not provide any advantage to generation task. The intuition we suggest is that predicting AoT requires extracting features of whole frames(videos) regarding time while the other tasks require extracting features of individual frames(images).

Furthermore, we explore categorical video generation with recent techniques in image generation, namely conditional batch normalization, spectral normalization, projection discriminator~\cite{miyato2018cgans}, and diversity-sensitive regularizer~\cite{mao2019mode}. They go along with our ArrowGAN, further improving the performance to achieve state-of-the-art video generation on UCF-101 as demonstrated by section \ref{sec:conditional}, and \ref{sec:categorical arrowgan}.

In experiments, we verify that applying Arrow-D improves the performance of three existing video-GANs (TGAN, VGAN, MoCoGAN) on Weizmann, UCFsports, and UCF-101 in terms of both quantitative and qualitative evaluation. We use video-version of inception score(IS)~\cite{salimans2016improved}, and Fr\'echet inception distance (FID)~\cite{heusel2017gans,Unterthiner2018TowardsAG} for evaluating the reality of generated samples. Also, section~\ref{sec:temporal self-supervision} demonstrates the superiority of learning AoT over state-of-the-art self-supervision technique on image and video domain, and section~\ref{sec:categorical arrowgan} shows that each component of categorical \kb{ArrowGAN} improves performance even with temporal inductive bias.
All codes and pre-trained weights will be publicly available for the research community. 

\begin{figure*}
\centering
\includegraphics[width=1.0\textwidth]{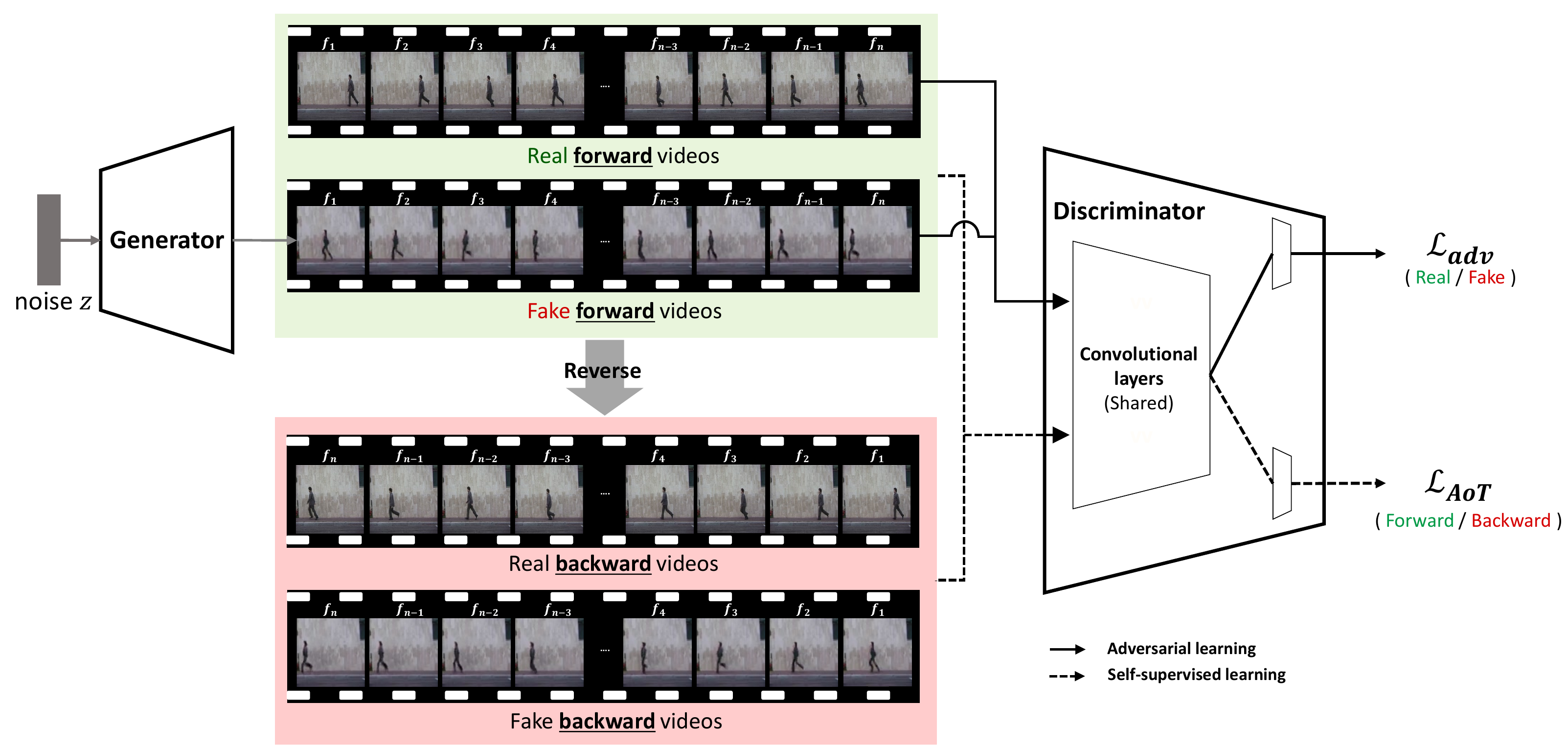}
\caption{Overview of ArrowGAN framework. The solid lines indicate the adversarial inputs, and the dashed lines indicate the self-supervised inputs by reversal for predicting Arrow of Time(AoT). The discriminator learns to classify real forward and backward videos while the generator learns to synthesize forward videos. 
}
\vspace{-1mm}
\label{fig:networks}
\end{figure*}

\section{Related work}
\subsection{Video generation}
Recently, many studies have extended the image-level generative models to video generation. VideoVAE~\cite{he2018probabilistic} propose the recurrent Variational Auto-Encoder(VAE) for conditional video generation. Also, \cite{aich2020non} synthesize videos with learned latent space via non-adversarial learning. Furthermore, generative adversarial networks (GANs) have been employed to generating videos by designing generators specialized for handling temporal coherency of videos. VGAN~\cite{vondrick2016generating} defines videos as a linear combination of dynamic foreground and static background and trains a two-stream generator against a spatio-temporal discriminator. TGAN~\cite{saito2017temporal} splits the generator into a temporal generator and an image generator. The temporal generator maps a primary latent vector to temporally varying latent vectors for individual frames, and the image generator produces each frame. MoCoGAN~\cite{tulyakov2018mocogan} uses the gated recurrent unit (GRU)~\cite{chung2014empirical} to further model the temporal relationship between latent vectors of different frames. On the other hand, we guide the discriminator to learn extra knowledge about the physical world which works as an inductive bias for the generator to produce more realistic videos.

\kb{Recently, Temporal GAN v2 (TGANv2)~\cite{Saito2020TrainSG} focuses on a training technique based on understanding the relationship between the resolution of the generated images and the computational cost. They introduce a stack of small sub-generators and specific training strategy with sub-sampling technique multiple times in the generator based on the architecture of TGAN~\cite{saito2017temporal}. In addition, DVD-GAN~\cite{Clark2019AdversarialVG} exploits the architecture of BigGAN~\cite{brock2018large} based on MoCoGAN~\cite{tulyakov2018mocogan} for large batch size. Though they achieve great performance in video generation, DVD-GAN requires a huge computational cost. In this paper, we focus on the effectiveness of combination of video-GANs and self-supervisory task without additional annotation. Since recent studies also rely on conceptual structure of the aforementioned models, there is no dramatic change in structure. Therefore, we set three models~\cite{vondrick2016generating,saito2017temporal,tulyakov2018mocogan} as our baselines.}

For \textit{categorical} video generation, TGAN~\cite{saito2017temporal} concatenates class labels to inputs for the generator and the discriminator~\cite{mirza2014conditional}, while MoCoGAN
~\cite{tulyakov2018mocogan} concatenates a condition labels to GRU unit and uses an auxiliary classifier \cite{odena2017conditional}. Our model employs more recent techniques, \eg, conditional batch normalization~\cite{de2017modulating} and projection discriminator~\cite{miyato2018cgans} which are shown effective in image generation.

Video prediction methods also aim to generate future frames in a given environment~(\eg{initial frames, semantic map, or reference videos}). \cite{Ranzato2014VideoM,Finn2016UnsupervisedLF} propose the deterministic modes for generating videos with initial frames. In addition, \cite{babaeizadeh2018stochastic,denton2018stochastic} utilize stochastic variation models that could learn not only fixed prior but learned prior distribution. Also,~\cite{Wang2018VideotoVideoS,pan2019video,Chan2019EverybodyDN} produce frames through translation techniques from reference frames to target videos. However, predicted videos are little free from initial cues. On the other hand, we focus on video generation from only latent code without such dense hints.

\subsection{Self-supervised learning and arrow of time}
Self-supervised learning has been shown useful learning discriminative feature representation for downstream tasks without expensive human supervision. For images, a wide range of pretext tasks exists, namely, reconstructing images~\cite{rumelhart1985learning,Vincent2008ExtractingAC} and predicting pseudo-labels (relative localization~\cite{doersch2015unsupervised}, and rotation prediction~\cite{gidaris2018unsupervised}). Especially, rotation prediction is combined with a discriminator in GAN framework to improve image generation performance \cite{chen2019self}.

In video analysis, self-supervision from time has been shown effective for various tasks. Order prediction network \cite{lee2017unsupervised} is pre-trained by ordering shuffled frames in videos to improve action recognition. Also, per-frame embeddings learned by temporal cycle-consistency~\cite{dwibedi2019temporal} can be used for video alignment (dense temporal correspondence) between video pairs and anomaly detection. Note that these methods extract frame-wise features to predict pretext tasks handling multiple frames.

\textit{Arrow of Time}(AoT) denotes a direction of motion in videos, \ie, forward or backward. Traditional handcrafted features for video analysis are shown to be able to tell AoT~\cite{pickup2014seeing}. In the modern machine learning era, classifying AoT is employed as a self-supervision task for convolutional networks to learn video representation~\cite{wei2018learning}. The learned embeddings of videos are useful for downstream tasks such as action recognition. In this paper, we show that combining AoT with video-GANs improves \kb{its} performance while frame-level self-supervision methods (predicting rotation and ordering) do not.

\begin{table*}[width=1.0\textwidth, pos=t]
\caption{Arrow-D on VGAN~\cite{vondrick2016generating}}
\centering\resizebox{1.0\textwidth}{!}{
\begin{tabular}{llll}
\hline
Module                  & Layers                                     & Input size                   & Output size                  \\ \hline
Input videos            & -                                          & T x 3 x 64 x 64              & -                            \\ \hline
\multirow{6}{*}{\begin{tabular}[l]{@{}l@{}}Video feature extractor\\(shared)\end{tabular}}& Noise                                      & T x 3 x 64 x 64              & T x 3 x 64 x 64              \\
                        & Conv3d(4, 2, 1), Noise, LeakyReLU(0.2)     & T x 3 x 64 x 64              & (Tx1/2) x 128 x 32 x 32      \\
                        & Conv3d(4, 2, 1), BN, Noise, LeakyReLU(0.2) & (Tx1/2) x 128 x 32 x 32      & (Tx1/4) x 256 x 16 x 16      \\
                        & Conv3d(4, 2, 1), BN, Noise, LeakyReLU(0.2) & (Tx1/4) x 256 x 16 x 16      & (Tx1/8) x 512 x 8 x 8        \\
                        & Conv3d(4, 2, 1), BN, Noise, LeakyReLU(0.2) & (Tx1/8) x 512 x 8 x 8        & (Tx1/16) x 1024 x 4 x 4      \\
                        & Flatten                                    & (Tx1/16) x 1024 x 4 x 4      & {[}1024 x(Tx1/16)x4x4{]} x 1 \\ \hline\hline
GAN Discriminator       & Linear({[}1024 x(Tx1/16)x4x4{]}, 1)            & {[}1024 x(Tx1/16)x4x4{]} x 1~~& 1 x 1                        \\ \hline
AoT Classifier          & Linear({[}1024 x(Tx1/16)x4x4{]}, 1)            & {[}1024 x(Tx1/16)x4x4{]} x 1~~& 1 x 1                        \\ \hline
\end{tabular}}
\label{tab:ArrowD_VGAN}
\end{table*}
\begin{table*}[width=1.0\textwidth, pos=t]
\caption{Arrow-D on TGAN~\cite{saito2017temporal}}
\centering\resizebox{1.0\textwidth}{!}{
\begin{tabular}{llll}
\hline
Module                                   & Layers                              & Input size              & Output size             \\ \hline
Input video                              & -                                   & T x 3 x 64 x 64         & -                       \\ \hline
\multirow{3}{*}{\begin{tabular}[l]{@{}l@{}}Video feature extractor\\(shared)\end{tabular}} & Conv3d(4, 2, 1), LeakyReLU(0.2)     & T x 3 x 64 x 64         & (Tx1/2) x 64 x 32 x 32  \\
                                         & Conv3d(4, 2, 1), BN, LeakyReLU(0.2) & (Tx1/2) x 64 x 32 x 32  & (Tx1/4) x 128 x 16 x 16 \\
                                         & Conv3d(4, 2, 1), BN, LeakyReLU(0.2) & (Tx1/4) x 128 x 16 x 16 & (Tx1/8) x 256 x 8 x 8   \\
                                         & Conv3d(4, 2, 1), BN, LeakyReLU(0.2) & (Tx1/8) x 256 x 8 x 8   & (Tx1/16) x 512 x 4 x 4  \\ \hline\hline
GAN Discriminator                                  & Conv2d(4, 1, 0)                     & (Tx1/16) x 512 x 4 x 4  & (Tx1/16) x 1            \\ \hline
AoT Classifier                      & Conv2d(4, 1, 0)         & (Tx1/16) x 512 x 4 x 4  & (Tx1/16) x 1            \\ \hline
\end{tabular}}
\label{tab:ArrowD_TGAN}
\end{table*}

\section{Method}
\subsection{ArrowGAN}
Humans can easily distinguish videos that are running forwards or backwards thanks to general understanding about a fundamental aspect of our life; time. Our intuition is to teach discriminators and generators to have a sense of \emph{Arrow of Time} (AoT) as humans do. Therefore, we aim to equip existing video-GANs with the sense of time to build ArrowGAN. We first introduce classifying AoT as an auxiliary task to the discriminators and explain how the generators benefit from it.

\subsubsection{Arrow discriminator.}
We first devise our discriminator $D$ to distinguish real and fake videos as well as forward and backward videos, similar to ACGAN~\cite{odena2017conditional}. $D$ outputs likelihood of an input video $x$ being real, $p(\text{real}|x )$, and likelihood of being forward $p(\text{forward} |x )$ at the same time. $D$ consists of shared convolutional layers and two branches for each output, respectively. 
For a given video $x$, we can easily obtain a pair of a forward video $x_{\text{foward}}$ and a backward video $x_{\text{backward}}$ by reversing it, to design self-supervised binary cross entropy loss:

\begin{equation}
    \centering
    \mathcal{L}_\text{AoT}^{D} = - \sum_{a \in A}\mathbb{E}_{x_{a} \sim  p_{\text{data}}} [ \log p(a|x_{a}) ] 
    \label{eq:d_arrow}
\end{equation}
where $A=\{\text{forward, backward}\}$ is set of AoTs,~$p_{\text{data}}$ denotes distribution of real videos. \kb{Please, note that our Arrow-$D$ calculate the \textbf{AoT loss} only on both forward-running and backward-running real videos (not on the generated videos). And AoT loss for $G$ is calculated on both forward-running and backward-running generated videos.} By minimizing above objective, the discriminator learns to classify AoT.

Meanwhile, we compute the adversarial loss:
\begin{equation}
    \mathcal{L}_{adv} =  \mathbb{E}_{x \sim  p_{data}}  [ \log D(x)  ]
    \\ + \mathbb{E}_{z \sim  p_z}  [ \log (1- D(G(z))  ]
    \label{eq:adv}
\end{equation}
and $D$ tries to maximize it, distinguishing real and generated videos. \kb{We note that Arrow-$D$ calculate the \textbf{adversarial loss} only on forward-running real videos and forward-running generated videos (not on both backward-running real videos and backward-running generated videos). Likewise, the adversarial loss for $G$ is calculated only on forward-running generated videos.}

Thus the full objective for $D$ becomes
\begin{equation}
    \mathcal{L}^{D} = - \mathcal{L}_{adv} + \alpha\mathcal{L}^{D}_\text{AoT}
    \label{eq:d_full}
\end{equation}
where $\alpha$ is a hyperparameter for controlling importance between two terms.

We train the two branches of the discriminator with their respective inputs as shown in \Fref{fig:networks}. The adversarial branch receives real and fake videos to compute ${\mathcal{L}_{adv}}$\kb{~(\ie{, solid line})}. Besides, the AoT branch receives forward and backward videos of both real and fake videos to compute ${\mathcal{L}^{D}_{\text{AoT}}}$\kb{~(\ie{, dashed line})}. By doing so, the shared convolutions learn features for realistic and forward-running videos from both tasks.

\subsubsection{Arrow generator.}
We expect our generator $G$ to synthesize videos with motions running forward in time as an inductive bias. To this end, $G$ tries to maximize the probability of generated videos being classified as forward by $D$, formulated as below:
\begin{equation}
    \mathcal{L}^{G}_\text{AoT} = - \sum_{a \in A}\mathbb{E}_{z \sim p_z} [ \log p(a|G(z)_{a}) ],
    \label{eq:g_arrow}
\end{equation}
where $G(z)$ is the generated video from noise $z$. By minimizing above objective, the generator learns to generate forward videos.
We note that $G$ only produces forward videos, but $D$ receives both forward and backward generated videos by reversal, to encourage them to resemble real videos.
Meanwhile, the generator tries to minimize the adversarial loss $\mathcal{L}_{adv}$ (\Eref{eq:adv}) for realism.

By minimizing full objective for $G$,
\begin{equation}
    \mathcal{L}^{G} = \mathcal{L}_{adv} + \beta\mathcal{L}^{G}_\text{AoT}
    \label{eq:g_full}
\end{equation}
$G$ learns to generate realistic as well as forward-running videos, 
where $\beta$ is a hyperparameter for balancing importance between two terms.

\begin{table*}[width=1.0\textwidth, pos=t]
\caption{Arrow-D on MoCoGAN~\cite{tulyakov2018mocogan}}
\centering
\resizebox{1.0\textwidth}{!}{
\begin{tabular}{llll}
\hline
Module                  & Layers                                                 & Input size            & Output size           \\ \hline
Input videos            & -                                                      & T x 3 x 64 x 64       & -                     \\ \hline
\multirow{3}{*}{\begin{tabular}[l]{@{}l@{}}Video feature extractor\\(shared)\end{tabular}}& Noise, Conv3d(4, (1,2,2), (0,1,1)), LeakyReLU(0.2)     & T x 3 x 64 x 64       & (T - 3) x 64 x 32 x 32  \\
                        & Noise, Conv3d(4, (1,2,2), (0,1,1)), BN, LeakyReLU(0.2) & (T - 3) x 64 x 32 x 32  & (T - 6) x 128 x 16 x 16 \\
                        & Noise, Conv3d(4, (1,2,2), (0,1,1)), BN, LeakyReLU(0.2) & (T - 6) x 128 x 16 x 16 & (T - 9) x 256 x 8 x 8   \\ \hline\hline
GAN Discriminator       & Conv3d(4, (1, 2, 2), (0, 1, 1))                        & (T - 9) x 256 x 8 x 8   & (T - 12) x 1 x 4 x 4    \\ \hline
AoT Classifier          & Conv3d(4, (1, 2, 2), (0, 1, 1))                        & (T - 9) x 256 x 8 x 8   & (T - 12) x 1 x 4 x 4    \\ \hline
\end{tabular}}
\label{tab:ArrowD_MoCoGAN}
\end{table*}
\subsubsection{Application on the baselines.}
We apply our ArrowGAN framework on three baselines: VGAN~\cite{vondrick2016generating}, TGAN~\cite{saito2017temporal} and MoCoGAN~\cite{tulyakov2018mocogan}. For all instances, we augment their discriminators with an auxiliary AoT classifier \kb{and we leave their generator without any modification. Afterwards, we }train the discriminators and generators with AoT losses ($\mathcal{L}^{D}_\text{AoT}$ and $\mathcal{L}^{G}_\text{AoT}$). While VGAN and MoCoGAN fits to our formulation, we use wasserstein distances~\cite{Arjovsky2017WassersteinGA} for adversarial losses of TGAN following their implementation. Image discriminator in MoCoGAN is left intact.
In all experiments, $(\alpha,~\beta)$ are set to ($1.0,~0.2$) on Weizmann and UCFsports, and ($1.0,~0.5$) on UCF-101 respectively for considering the complexity of datasets.

\subsubsection{Implementation details.}
\kb{We describe more detailed architectures of our Arrow Discriminator (Arrow-D) in Table~\ref{tab:ArrowD_VGAN},~\ref{tab:ArrowD_TGAN}, and~\ref{tab:ArrowD_MoCoGAN} that inform  hyperparameters of each Arrow-D on VGAN~\cite{vondrick2016generating}, TGAN~\cite{saito2017temporal}, and MoCoGAN~\cite{tulyakov2018mocogan} respectively. In all tables, T denotes the length of frames. Conv3d(K, S, P) means that 3D convolution with the kernel size of K, the stride of S, and the padding of P. And, BN denotes the batch normalization, LeakyReLU(R) is the activation function of leaky rectified linear unit with the negative slope of R. We set the value of R as 0.2 in our all experiments. Also, in all tables, the video feature extractor is shared, and rows below it indicate each branch: adversarial learning and self-supervised learning.}

\kb{We train our networks using Adam optimizer~\cite{kingma2014adam} with hyper-parameters set to $\alpha=0.0002, \beta_{1}=0.5, \beta_{2}=0.999$. In particular, Arrow-D on VGAN and MoCoGAN have a noise layer for more stable training. Noise layer adds a Gaussian noise of the same size to features extracted from each of the convolutional layers. We set the weight of noise to 0.1. Also Arrow-D on MoCoGAN is based on the PatchGAN~\cite{isola2017image}. We note that we have used generators without any modification.}

\subsection{Extension to categorical generation}
\label{sec:conditional}

In this section, we further explore categorical generation, \ie, video generation given class labels. We first describe how to adapt techniques from image-GANs for video-GANs. Then we formulate categorical ArrowGAN.

\subsubsection{Closing gap vs. image generation.}
First, we employ \kb{\textit{conditional batch normalization}}~(CBN)~\cite{dumoulin2017learned,de2017modulating} to inject class labels into the generator. CBN layer modulates feature activations in the generator by manipulating parameters for batch normalization to condition output images on the class labels. Here, we insert CBN layer before all deconvolutions for video generation. 

Second, we adopt the \kb{\textit{projection discriminator}} for class scores~\cite{miyato2018cgans}. The projection discriminator computes inner product between its penultimate feature vector and class embedding, and regards the inner product as score for the class. It can be naturally extended to 3D convolutions for video discriminator as below:

\begin{equation}
\begin{gathered}
        D_{\text{proj}}(x, y) = y^{T} V \Phi (x) + \psi (\Phi ( x ))
\end{gathered}
\label{eq:projection}
\end{equation}
where $y$ is an one-hot vector denoting class label, $V$ is class embedding matrix, $\Phi$ extracts feature vector at the penultimate layer of $D$ through 3D convolutions, and $\psi$ is a fully-connected layer producing a scalar score for realism. 

Third, we exploit the \kb{\textit{spectral normalization layer}}~\cite{miyato2018spectral} on both video and frame discriminators for stable training. 

Lastly, we enforce the diversity in video and alleviate mode collapse problem by explicitly promoting the distance between generated videos over the distance between latent vectors~\cite{yang2018diversitysensitive}. Specifically, we minimize its reciprocal:
\begin{equation}
\begin{gathered}
        \mathcal{L}_{\text{div}} = \frac{d(z_1, z_2)} {d (G(z_1,y) , G(z_2,y))}
\end{gathered}
\label{eq:div}
\end{equation}
where $d$ denote L1 distance between latent vectors(\ie{, $z_{1}, z_{2}$}) and the distance between generated videos(\ie{, $G(z_{n}, y)$}) respectively.

\begin{figure*}
\centering
\includegraphics[width=1.0\textwidth]{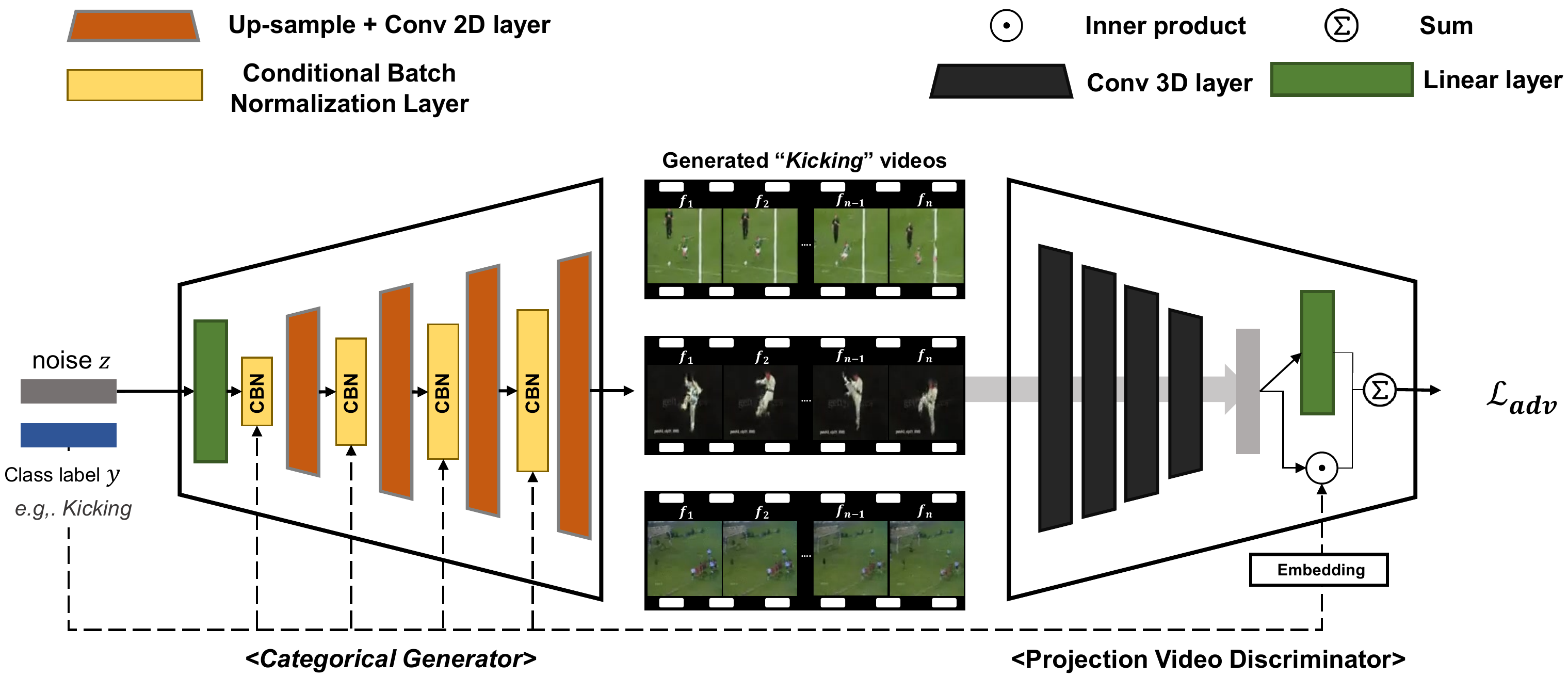}
\caption{Overview of categorical ArrowGAN framework. Our categorical generator consists of up-sample convolutional layers and conditional batch normalization(CBN) layers. Video discriminator project a extracted video feature and embedded class-label into shared space for calculating adversarial outputs. We omit the diversity sensitive part in this figure.}
\label{fig:categorical_networks}
\end{figure*}

\subsubsection{Categorical ArrowGAN.}
We train our categorical ArrowGAN with below objective functions: \textit{hinge} version of the adversarial losses~\cite{Zhang2018SelfAttentionGA,miyato2018cgans}, AoT loss and diversity-sensitive regularizer. 

Training of $G$ and $D$ is achieved via solving min-max problem given by

\begin{equation}
\begin{split}
    \mathcal{L}^{D}_\text{adv}=  &\mathbb{E}_{x, y\sim  P_{data}}  [\min(0,1+~D_{proj}(x,y))  ]~+\\
     &\mathbb{E}_{\tilde{z}\sim  p_z, y\sim  P_{data}}  [ \min (0,1-D_{proj}(G(z,y),y))  ]
     \\
     \mathcal{L}^G_\text{adv} =  &- \mathbb{E}_{\tilde{z}\sim  p_z, y\sim  P_{data}}  [D(G(z,y),y)  ] 
     \\
\label{eq:vgan}
\end{split}
\end{equation}

Finally, the full objectives of the discriminators and the generator are formulated as:
\begin{equation}
\begin{split}
     \mathcal{L}^{D} &= -\mathcal{L}^{D}_\text{adv} + \lambda_{1}{\mathcal{L}}^{D}_\text{AoT}\\
     \mathcal{L}^{G} &= \mathcal{L}^{G}_\text{adv} + \lambda_{2}\mathcal{L}^{G}_\text{AoT} + \lambda_{3}\mathcal{L}_\text{div}
\end{split}
\label{eq:generator objective function}
\end{equation}
where $\lambda_{1}$, $\lambda_{2}$, and $\lambda_{3}$ are set to 1, 0.5 and 0.2, respectively. For MoCoGAN, we keep using their image discriminator for below losses:
\begin{equation}
\begin{split}
     \mathcal{L}_\text{frame}~ &= \mathbb{E}_{x_{i}\sim  p_\text{data}}  [ \log D_\text{frame}(x_{i})  ]  +\\ &~~~~~~~~~~~~~~~~~~~~~~~~~~~\mathbb{E}_{\tilde{x_{i}}\sim  P_{G}} [ \log (1-D_\text{frame}(\tilde{x_{i}})) ] \\
    \mathcal{L}^{D_\text{frame}} &= -\mathcal{L}_\text{frame} \\
     \mathcal{L}^{G_\text{moco}} &= \mathcal{L}^{G} + \mathcal{L}_\text{frame}
\end{split}
\label{eq:image}
\end{equation}
where $x_{i}$ and $\tilde{x_{i}}$ are $i$-th frame from real and generated videos respectively.

\section{Experiments}
In this section, we describe experimental setups and conduct a set of experiments to show superiority of AoT over rotation, effectiveness of Arrow-D on three baselines, and improved performance on categorical generation.

In particular, Arrow-D on VGAN and MoCoGAN have a noise layer for more stable training. Noise layer adds a Gaussian noise of the same size to features extracted from each of the convolutional layers. we set the weight of noise to 0.1. We note that we have used generators without any modification.

Our code is implemented with PyTorch~\cite{paszke2017automatic} and our model is trained on a single GTX 2080Ti. The code and the trained model will be publicly available online.

\subsubsection{Datasets.}  We use three datasets: Weizmann Action, UCFsports, and UCF-101.

\begin{itemize}
\item \textit{Weizmann Action dataset} \footnote{\url{http://www.wisdom.weizmann.ac.il/~vision/SpaceTimeActions.html}}~\cite{blank2005actions} contains 90 videos of ten action categories \eg, \textit{jump, run}, etc., for nine people. In our experiment, we used nine action categories except \textit{skip} following the baselines.

\item \textit{UCFsports}\footnote{\url{https://www.crcv.ucf.edu/data/UCF_Sports_Action.php}}~\cite{soomro2014action} is a set of various actions collected from ten classes of sports, \eg, \textit{diving, swing-bench}, etc., from TV-broadcasting programs. It has 150 videos in total. 
We clarify that UCFsports is not a subset of the UCF-101 dataset.

\item \textit{UCF-101}\footnote{\url{https://www.crcv.ucf.edu/data/UCF-101.php}} \cite{soomro2012UCF101} is a large-scale benchmark video dataset that contains 13,320 videos of 101 different sport categories. Videos are collected from Youtube.
\end{itemize}

In unconditional video synthesis, we resized each frame to 85x64 and center-cropped to prepare 64x64 videos. In categorical video synthesis, we resized all frames to 96x96 without any crop. The above pre-processings match with the respective baseline methods for fair comparison.

\subsubsection{Evaluation metrics.} We measure \emph{video-version} of inception score (IS) and Fr\'echet inception distance (FID) for quantitative evaluation.
The original IS~\cite{salimans2016improved} measures KL divergence between conditional label distribution and marginal class distribution :
\begin{equation}
    \exp[\mathbb{E}_{z \sim P_z}[d_{KL}(p(y|G(z))~,~p(y))]]
\label{eq:IS}
\end{equation}
computed by Inception models such as Inception-V3~\cite{szegedy2016rethinking}.
The original FID~\cite{heusel2017gans} computes 2-Wasserstein distance between real and generated data:
\begin{equation}
\begin{split}
    ||\mu_{x} - \mu_{g}||^{2}_{2} + \mathrm{Tr}(\Sigma_{x} + \Sigma_{g} - 2(\Sigma_{x}\Sigma_{g})^{\frac{1}{2}})
\end{split}
\label{eq:FID}
\end{equation}
where $(\mu_{x},\Sigma_{x}), (\mu_{g},\Sigma_{g})$ are mean and co-variance of extracted features from Inception models, given real and generated data. IS and FID evaluate general performance of generative models and FID can detect intra-class mode dropping, and is more similar to human judgement than IS~\cite{heusel2017gans,lucic2018gans}.

For evaluating \emph{video-}IS, we used code and pre-trained\footnote{Pre-trained C3D model on UCF-101 can be downloaded from~\url{https://www.dropbox.com/s/bf5z2jw1pg07c9n/c3d_resnet18_ucf101_r2_ft_iter_20000.caffemodel?dl=0}} C3D model~\cite{tran2015learning} provided by authors of TGAN~\cite{saito2017temporal} 
We omit `video-' for brevity hereafter.
Additionally, we fine-tuned the C3D model on Weizmann and UCFsports, for each case.
For computing \emph{video-}FID, we used \textit{avgpool} feature vectors from \textit{mixed-5c} layer of I3D model fine-tuned on Kinetics dataset~\cite{carreira2017quo} following Fr\'echet video distance (FVD)~\cite{Unterthiner2018TowardsAG}. We name the video-FID as FVD in this paper. Particularly, we generated 10,000 samples videos from 4 random seeds and measured IS and FVD as in TGAN.

\subsection{The effectiveness of learning AoT}
We evaluate the effectiveness of ArrowGAN framework on three existing unconditional video-GANs: VGAN~\cite{vondrick2016generating}, TGAN~\cite{saito2017temporal} and MoCoGAN~\cite{tulyakov2018mocogan}, and a conditional video-GANs: Categorical MoCoGAN. We implemented the baselines in PyTorch~\cite{paszke2017automatic} and also report their inception scores (IS) alongside values taken from MoCoGAN paper.

\begin{table*}[width=1.0\textwidth]
\caption{Quantitative results of ArrowGANs in video-IS and Fr\'chet video distance(FVD). ArrowGAN framework improves all three baselines on all datasets. $\dagger$ notices that IS values are taken from MoCoGAN~\cite{tulyakov2018mocogan}, and other values are taken from our reproduced models.}
\centering\resizebox{.97\textwidth}{!}{
\begin{tabular}{c|cc|cc|cc}
\cline{1-7}\cline{1-7}
\multicolumn{1}{c|}{\multirow{2}{*}{Methods}} & \multicolumn{2}{c|}{\textit{Weizmann}} & \multicolumn{2}{c|}{\textit{UCFsports}} & \multicolumn{2}{c}{\textit{UCF-101}} \\ \cline{2-7}
&\multicolumn{1}{c}{IS($\uparrow$)} & \multicolumn{1}{c|}{~~~FVD($\downarrow$)~~~} & \multicolumn{1}{c}{IS($\uparrow$)} & \multicolumn{1}{c|}{~~~FVD($\downarrow$)~~~} & \multicolumn{1}{c}{IS($\uparrow$)} & \multicolumn{1}{c}{~~~FVD($\downarrow$)~~~}\\ \hline \hline

{VGAN~\cite{vondrick2016generating}} & 2.30$\pm$.01 & 46.91$\pm$.32 & 1.22$\pm$.00 & 100.16$\pm$.71 & 8.63$\pm$.01 / 8.18$\pm.05^{\dagger}$ & 85.64$\pm$.06\\ 
{Arrow-VGAN} & \textbf{3.37}$\pm$.01 & \textbf{40.99}$\pm$.36 & \textbf{2.52}$\pm$.02 & \textbf{93.25}$\pm$2.5  & \textbf{10.58}$\pm$.08 & \textbf{83.46}$\pm$.09\\ 
\hline\hline
{TGAN~\cite{saito2017temporal}} & 3.58$\pm$.01 & 34.62$\pm$.33 & 2.39$\pm$.02 & 80.86$\pm$1.6  & 11.17$\pm$.02 / 11.85$\pm.07^{\dagger}$ & 35.70$\pm$.13\\ 
{Arrow-TGAN} & \textbf{4.33}$\pm$.02 & \textbf{27.89}$\pm$.17 & \textbf{2.80}$\pm$.02 & \textbf{76.91}$\pm$1.5  & \textbf{12.45}$\pm$.09 & \textbf{35.16}$\pm$.05\\  
\hline\hline
{MoCoGAN~\cite{tulyakov2018mocogan}} & 4.25$\pm$.01 & 31.23$\pm$.26 & 2.74$\pm$.02 & 68.47$\pm$1.6  & 11.29$\pm$.07 / 12.42$\pm.03^{\dagger}$ & 36.15$\pm$.09 \\
{Arrow-MoCoGAN} & \textbf{4.41}$\pm$.00 & \textbf{28.24}$\pm$.07 & \textbf{3.01}$\pm$.01 & \textbf{68.07}$\pm$.07  & \textbf{12.18}$\pm$.02 & \textbf{34.37}$\pm$.07\\ 
\hline\hline
\multicolumn{1}{c|}{Real dataset} & \multicolumn{1}{c}{5.64} &  \multicolumn{1}{c|}{6.02} & \multicolumn{1}{c}{6.10} & 16.55 & \multicolumn{1}{c}{34.55} & 0.68 \\
\cline{1-7}\cline{1-7}
\end{tabular}
}

\label{tab:effective_uncond}
\end{table*}

\begin{table*}[width=1.0\textwidth]
\caption{Quantitative results of ArrowGANs in video-IS and Fr\'chet video distance. ArrowGAN framework improves categorical baseline.}
\centering\resizebox{.97\textwidth}{!}{
\begin{tabular}{c|cc|cc|cc}
\cline{1-7}\cline{1-7}
\multicolumn{1}{c|}{\multirow{2}{*}{Methods}} & \multicolumn{2}{c|}{\textit{Weizmann}} & \multicolumn{2}{c|}{\textit{UCFsports}} & \multicolumn{2}{c}{\textit{UCF-101}} \\ \cline{2-7}
&\multicolumn{1}{c}{IS($\uparrow$)} & \multicolumn{1}{c|}{~~~FVD($\downarrow$)~~~} & \multicolumn{1}{c}{IS($\uparrow$)} & \multicolumn{1}{c|}{~~~FVD($\downarrow$)~~~} & \multicolumn{1}{c}{IS($\uparrow$)} & \multicolumn{1}{c}{~~~FVD($\downarrow$)~~~}\\ \hline \hline

{Categorical MoCoGAN~\cite{tulyakov2018mocogan}} & 4.41$\pm$.01 & 30.69$\pm$.06 & \textbf{3.21}$\pm$.01 & 85.88$\pm$1.3 & 12.32$\pm$.14 & 41.06$\pm$.05 \\
Arrow-Categorical MoCoGAN & \textbf{4.65}$\pm$.01 & \textbf{28.23}$\pm$.04 & 2.90$\pm$.01 & \textbf{83.48}$\pm$.32  & \textbf{14.33}$\pm$.06 & \textbf{40.34}$\pm$.05\\
\hline\hline
\multicolumn{1}{c|}{Real dataset} & \multicolumn{1}{c}{5.64} &  \multicolumn{1}{c|}{6.02} & \multicolumn{1}{c}{6.10} & 16.55 & \multicolumn{1}{c}{34.55} & 0.68 \\
\cline{1-7}\cline{1-7}
\end{tabular}
}

\label{tab:effective_cond}
\end{table*}

\begin{figure*}
\centering
\includegraphics[width=1.0\textwidth, keepaspectratio]{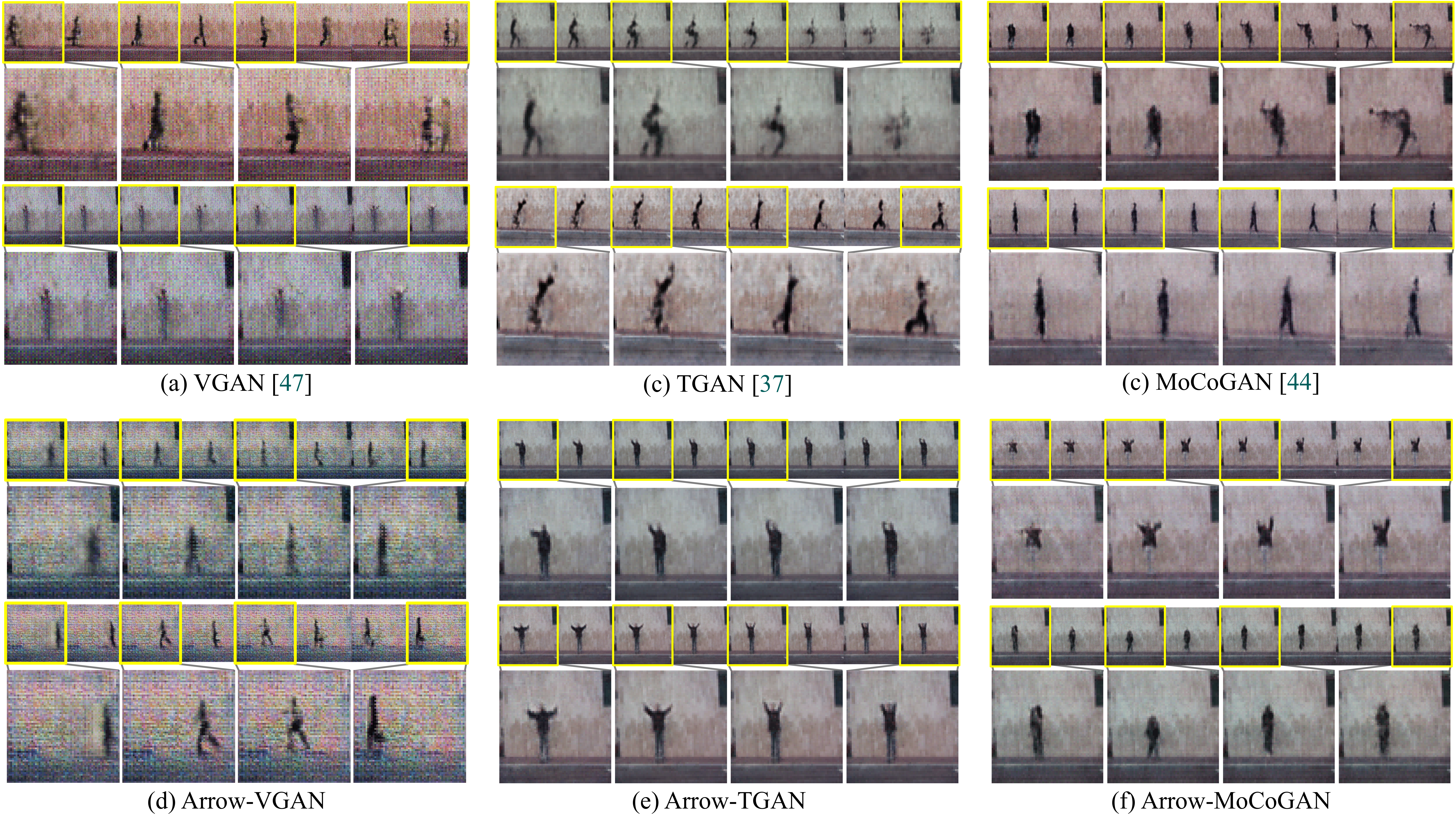}
\caption{Qualitative results of ArrowGAN. ArrowGAN generates recognizable action and human shapes than each baseline, especially temporal  changes and four limbs. Eight frames of each row go forward from left(first frame) to right(last frame). The yellow box depicts the zoomed frame.}
\label{fig:unconditional_qual}
\end{figure*}

\Tref{tab:effective_uncond} quantitatively shows consistent improvement of IS over all baselines and all datasets, when ArrowGAN is applied. Note that we do not modify their generators but only add learning AoT as an auxiliary task with minimal change on their discriminators. It implies that our ArrorGAN framework can be easily applied to other video-GANs for quality improvement. Notably, VGAN and TGAN achieve performance on par with their subsequent model only with proposed Arrow-D. Moreover, \Tref{tab:effective_cond} demonstrates the effectiveness of ArrowGAN framework with conditional video generation.

\Fref{fig:unconditional_qual} provides qualitative comparisons. \kb{In detail, the results of Arrow-VGAN and Arrow-TGAN show not only the shape of the action well alive, but the shape of the object could be recognized better than each baseline. Also, in the case of Arrow-MoCoGAN, the most realistic samples are generated. Consequentyl, we observe improvement for all cases, especially on fine details such as four limbs and temporal coherence. Note that we do not use explicit temporal coherence such as consistency between adjacent frames.}

\subsection{Comparison to other self-supervisory tasks}
\label{sec:temporal self-supervision}
We demonstrate that AoT is more beneficial to video-GANs than predicting rotation and temporal ordering of shuffled frames by an ablation study. We use MoCoGAN as a baseline, and augment it with different auxiliary self-supervisory tasks for comparison.

As a competitor, we consider predicting rotation~\cite{gidaris2018unsupervised,jing2018self} which is shown helpful for image generation task~\cite{chen2019self}. The input videos are randomly rotated by one among $\{0^{\degree}, 90^{\degree},$ $180^{\degree}, 270^{\degree}\}$, and we exploit the same training strategy of ArrowGAN. We impose predicting rotation of videos for video discriminator and predicting rotation of frames for frame discriminator of MoCoGAN. Because only MoCoGAN has a video discriminator and a frame discriminator. We also consider predicting whether frames are shuffled or not~\cite{Misra2016ShuffleAL} as another auxiliary task. We shuffle frames of each video randomly, and demand discriminator to verify whether inputs are shuffled or not.

\begin{table*}[width=1.0\textwidth]
\caption{Ablation study of self-supervision tasks. (a) indicates each baseline, and (b)$\sim$(f) indicates that self-supervisory tasks are adapted on (a), respectively. Predicting AoT(f) improves the baseline while verifying order(b), and predicting rotation(c-e) do not. The higher IS and the lower FVD mean better performance.}
\centering\resizebox{.97\textwidth}{!}{
\begin{tabular}{c|l|cc|cc}
\cline{1-6}\cline{1-6}
\multicolumn{1}{c|}{\multirow{2}{*}{Baseline}}& 
\multicolumn{1}{c|}{\multirow{2}{*}{Self-supervisory tasks}}
& \multicolumn{2}{c|}{\textit{Weizmann}} & \multicolumn{2}{c}{\textit{UCFsports}} \\ \cline{3-6}
& &\multicolumn{1}{c}{IS($\uparrow$)} & \multicolumn{1}{c|}{~~~FVD($\downarrow$)~~~} & \multicolumn{1}{c}{IS($\uparrow$)} & \multicolumn{1}{c}{~~~FVD($\downarrow$)~~~} \\ \hline
\hline
\multicolumn{1}{c|}{\multirow{6}{*}{\begin{tabular}[c]{@{}c@{}}MoCoGAN\\\cite{tulyakov2018mocogan}\end{tabular}}} &~~ (a) -~(baseline)&~~4.254$\pm$.01~~~& 31.23$\pm$.26 &~~2.735$\pm$.02~~~& 68.47$\pm$1.6\\
&~~ (b) Shuffled & 4.228$\pm$.01 & 34.57$\pm$.38 & 2.879$\pm$.01 & 70.19$\pm$1.1\\
&~~ (c) Rotation($D_{\textit{frame}}$) & 4.043$\pm$.01 & 33.03$\pm$.25 & 2.856$\pm$.01 & 75.93$\pm$.70 \\
&~~ (d) Rotation($D_{\textit{video}}$)& 4.269$\pm$.01 & 34.27$\pm$.11 & 2.771$\pm$.06 & 70.25$\pm$1.5 \\
&~~ (e) Rotation($D_{\textit{frame}}$~\&~$D_{\textit{video}}$)~~~& 4.186$\pm$.01 & 28.73$\pm$.34 & 2.851$\pm$.02 & 69.53$\pm$.92 \\
&~~ (f) \textbf{AoT(\textit{ours})}& \textbf{4.414}$\pm$.00 & \textbf{28.24}$\pm$.07 & \textbf{3.013}$\pm$.01 & \textbf{68.07}$\pm$.07  \\
\hline \hline
\multicolumn{2}{c|}{Real dataset} & \multicolumn{1}{c}{5.636} &  \multicolumn{1}{c|}{6.02} & \multicolumn{1}{c}{6.104} & 16.55 \\
\cline{1-6}\cline{1-6}
\end{tabular}
}
\label{tab:ablation_temporal}
\end{table*}

\Tref{tab:ablation_temporal} compares four self-supervisory tasks; predicting AoT, rotation of video, rotation of frame and verifying order. AoT improves all baselines the most, while other self-supervised techniques do not show meaningful improvement of the performance. It supports that AoT is a suitable auxiliary self-supervisory task for video-GANs. Rotation of frame have caused more performance degradation compared to other tasks of video-level~\ie, rotation of video, and shuffled as opposed to SSGAN~\cite{chen2019self}. We suggest that the property of target domain to be generated should be considered when choosing the self-supervisory task.

\begin{table}[width=0.99\linewidth]
\caption{Comparison with previous methods for categorical video generation. We achieve the state-of-the-art performance on UCF-101. IS of conditional TGAN is reported from~\cite{saito2017temporal}, and other values are taken from our implementations.}
\centering\resizebox{.99\linewidth}{!}{
\begin{tabular}{l|cc}
    \cline{1-3}\cline{1-3}
    \multirow{2}{*}{Methods} & \multicolumn{2}{c}{\textit{UCF-101}} \\\cline{2-3}
                                                  & IS($\uparrow$) &~~~FVD($\downarrow$)~~~ \\
    \hline\hline
    {conditional TGAN~\cite{saito2017temporal}} &~~~15.83$\pm$.18~~~&  - \\ 
    {categorical MoCoGAN~\cite{tulyakov2018mocogan}} &~~~12.32$\pm$.14~~~&41.06$\pm$.05 \\ \cline{1-3}
    {categorical ArrowGAN~(Ours)} &~~~\textbf{28.79}$\pm$.20~~~&\textbf{26.98}$\pm$.10\\
    \cline{1-3}\cline{1-3}
  \end{tabular}
}
\label{tab:conditional}
\end{table}

\begin{figure*}
\centering
\includegraphics[width=1.0\linewidth, keepaspectratio]{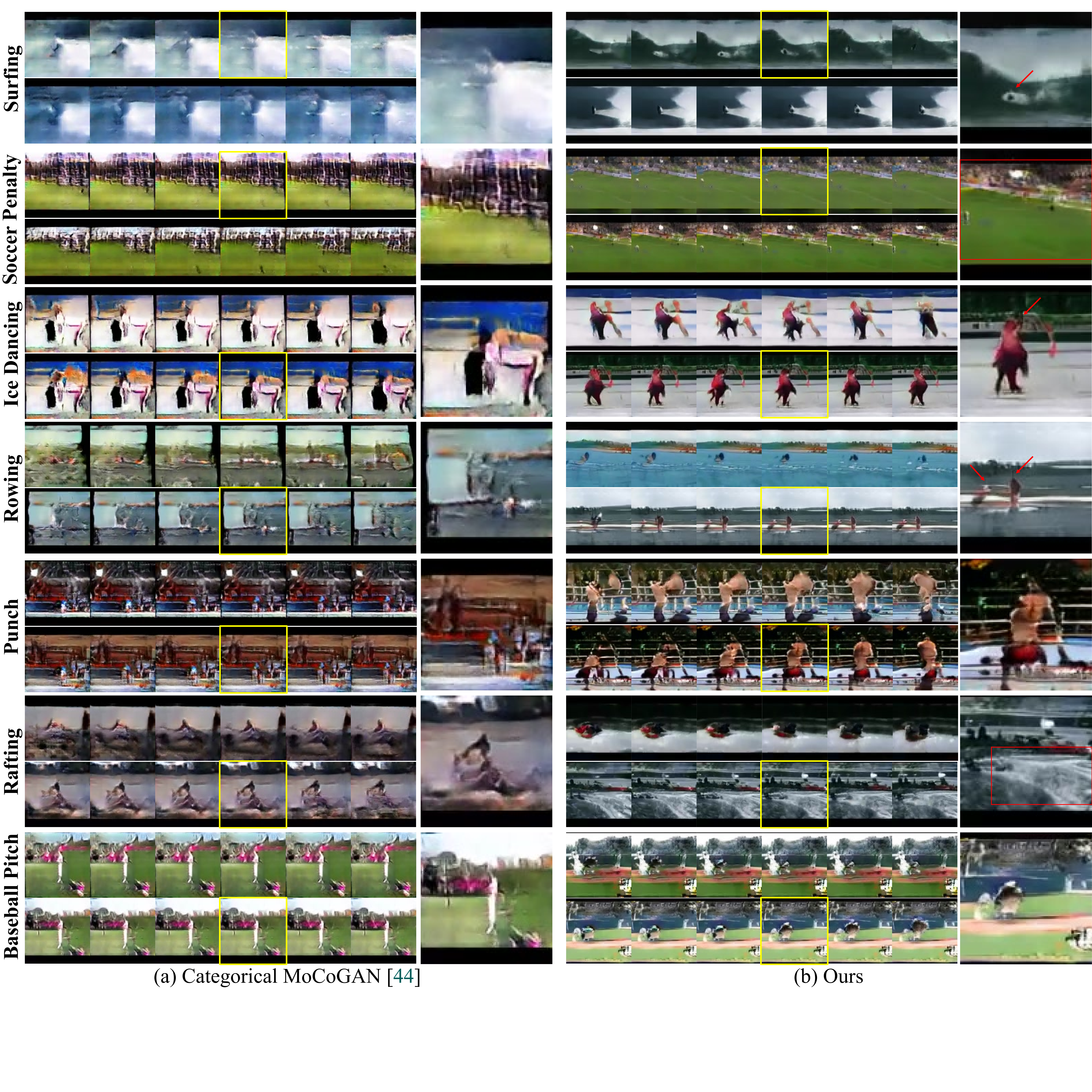}

\caption{Comparison with categorical MoCoGAN~\cite{tulyakov2018mocogan}. (a) Categorical MoCoGAN generates videos with little motion and hardly recognizable objects. (b) Categorical ArrowGAN produces more recognizable videos both in motion and objects. In detail, we can see skaters moving in ice dancing class, and boxers and fence in punch class. The yellow box depicts the zoomed frame for better qualitative comparison.}
\label{fig:conditional_results}
\end{figure*}

\subsection{Categorical ArrowGAN}
\label{sec:categorical arrowgan}

We demonstrate that categorical ArrowGAN outperforms existing categorical video generation methods (conditional TGAN and categorical MoCoGAN) on Weizmann, UCFsports and UCF-101.

\Tref{tab:conditional} shows that we yielded a boost of 12.96 of IS and 14.08 of FVD values than the state-of-the-art conditional TGAN and Categorical MoCoGAN quantitatively. In a next section, we analyze the effectiveness of each components of categorical ArrowGAN in detail.

\Fref{fig:teaser} demonstrates qualitative samples from categorical ArrowGAN trained on Weizmann, UCFsports, and UCF-101. \Fref{fig:teaser}(a,b) shows recognizable shapes and motions such as running, jumping, weightlifting and the balance. Moreover, we find that categorical ArrowGAN has the capability to generate on not only simple dataset, but a large-scale dataset, UCF-101 as shown in \Fref{fig:teaser}(c). In addition, our categorical ArrowGAN can generate videos with high resolution of 256x256 as shown in \Fref{fig:teaser}(d).

Also \Fref{fig:conditional_results} provides qualitative results of comparison with categorical MoCoGAN~\cite{tulyakov2018mocogan} on UCF-101. \kb{In case of surfing, rowing and ice dancing, the movement of actors~(red arrow) such as surfer can be recognized better than that of categorical MoCoGAN. In addition, our methods generate backgrounds~(red box) of behavior such as soccer field or water as well as foreground actions well. Generally, we note that categorical ArrowGAN generates more realistic videos with respect to dynamic action thanks to the sense of AoT.}

\begin{table*}[width=1.0\textwidth]
\vspace{-5mm}
\caption{Effectiveness of components to build the categorical ArrowGAN. First row indicates the  categorial MoCoGAN~\cite{tulyakov2018mocogan} as our baseline, and other rows shows the effectiveness of each component. The projection video discriminator, regularizer of diversity-sensitive, and our proposed self-supervisory task, AoT, contribute to improvement of IS and FVD scores.}
  \centering\resizebox{.91\textwidth}{!}{
  \begin{tabular}{ccc|cc|cc|cc}
    \cline{1-9}\cline{1-9}
    \multicolumn{3}{c|}{Methods} & \multicolumn{2}{c|}{\textit{Weizmann}} & \multicolumn{2}{c|}{\textit{UCFsports}} &
    \multicolumn{2}{c}{\textit{UCF-101}} \\ \cline{1-9}
    Projection&\begin{tabular}[c]{@{}c@{}}Diversity\\sensitive\end{tabular}&~~\begin{tabular}[c]{@{}c@{}}AoT
  \end{tabular}~~& IS($\uparrow$) &~~~FVD($\downarrow$)~~~& IS($\uparrow$) &~~~FVD($\downarrow$)~~~&IS($\uparrow$) &~~~FVD($\downarrow$)~~~\\
  \hline\hline
            &                   &               &~~~4.41$\pm$.01~~~     & 30.69$\pm$.06 &
            ~~3.21$\pm$.01~~&~~85.88$\pm$1.3~~&~~12.32$\pm$.14~~~& 41.06$\pm$.05 \\\hline
$\checkmark$&                   &               & 4.86$\pm$.01         & 21.62$\pm$.05 &~~3.26$\pm$.02~~&~~78.49$\pm$.49~~& 21.68$\pm$.27  & 28.02$\pm$.05\\
$\checkmark$&$\checkmark$&& 4.98$\pm$.01 & 15.90$\pm$.03 &~~3.15$\pm$.02~~&~~78.53$\pm$1.3~~& 22.91$\pm$.12& 28.93$\pm$.08 \\
$\checkmark$& $\checkmark$ & $\checkmark$& \textbf{5.35}$\pm$.03 & \textbf{15.87}$\pm$.03 &~~\textbf{3.57}$\pm$.01~~&~~\textbf{77.99}$\pm$.46~~& \textbf{28.79}$\pm$.20 & \textbf{26.98}$\pm$.10 \\
\hline \hline
\multicolumn{3}{c|}{Real dataset} & \multicolumn{1}{c}{5.64} & \multicolumn{1}{c|}{6.02} & \multicolumn{1}{c}{~~6.10~~} &~~16.55~~& \multicolumn{1}{c}{34.55} & 0.68 \\
\cline{1-9}\cline{1-9} 
\end{tabular}
}
\label{tab:ablation all}
\end{table*}
\subsubsection{Quantitative ablation study.}
In order to verify the effectiveness of individual components in categorical ArrowGAN, we start from categorical MoCoGAN~\cite{tulyakov2018mocogan} and add components step by step; conditional batch normalization plus projection discriminator with spectral normalization, diversity-sensitive regularizer, and AoT.

\begin{figure*}
\centering
\includegraphics[width=1.0\textwidth, keepaspectratio]{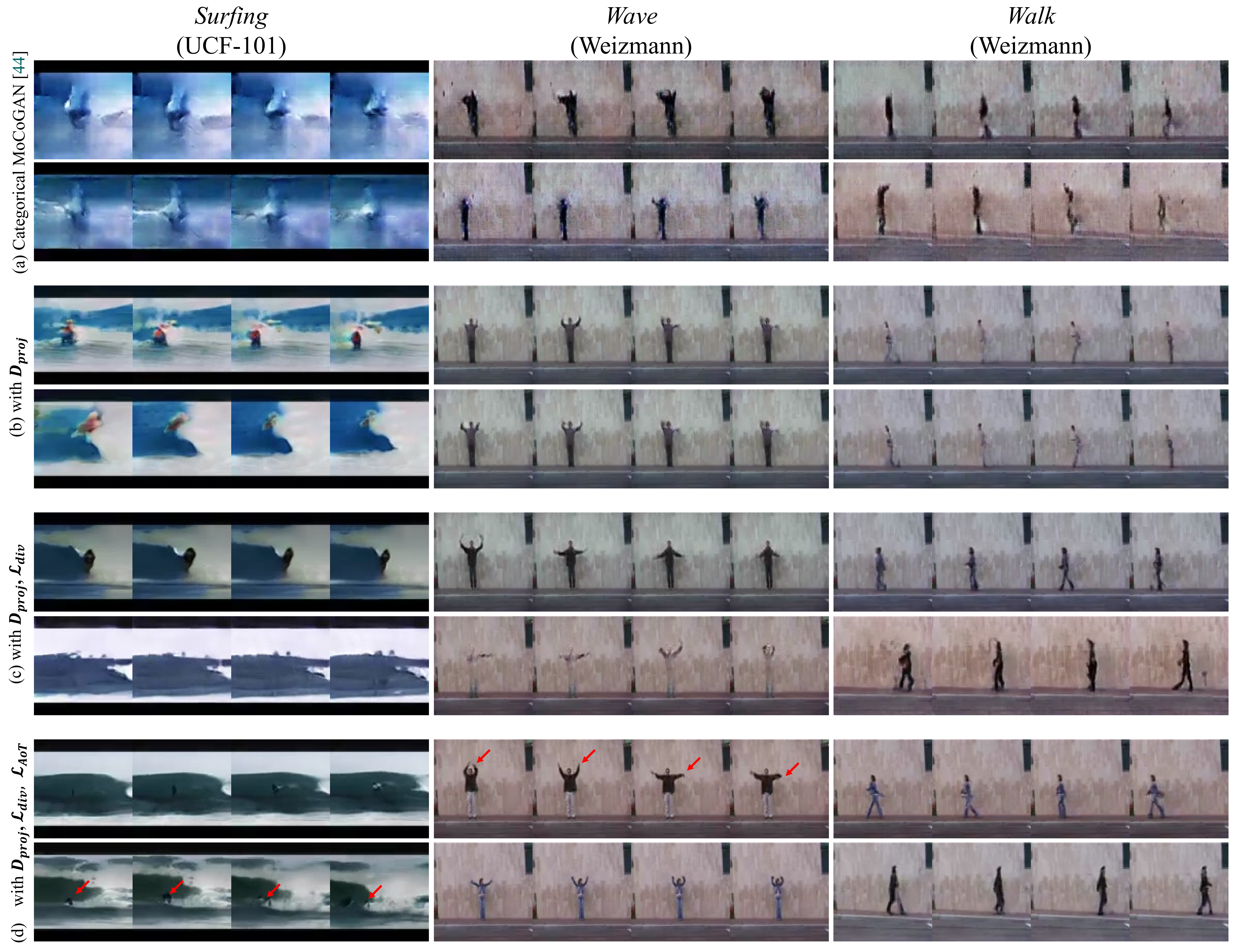}
\caption{Qualitative ablation study of categorical ArrowGAN~\cite{tulyakov2018mocogan} on Weizmann\&UCF-101. From (a) to (d), we add the projection discriminator($\mathcal{D}_{proj}$) and categorical batch normalization, diversity-sensitive regularizer($\mathcal{L}_{div}$), and AoT self-supervisory task($\mathcal{L}_{AoT}$) sequentially. Projection discriminator and $\mathcal{L}_{div}$ help to generate more various samples, and AoT helps to generate more realistic samples (a surfer and players appear, and more recognizable action can be detected in (d)).
}
\vspace{-1.7mm}
\label{fig:ablation}
\end{figure*}

\Tref{tab:ablation all} shows that all the components contribute to improvement of quality.
Starting from categorical MoCoGAN, we replace concat-generator and ac-discirminator by conditional batch normalization (CBN) and projection discriminator. The first \& second rows show that we gain the improvement of IS with 9.36 and FVD with 13.04 on UCF-101 via constructing the robust baseline with the projection discriminator.

Next, we conduct the comparison by adding the diversity-sensitive regularizer and AoT discriminator respectively on the preceding projection baseline. The model adopted only with diversity-sensitive regularizer gains good performance of IS and FVD in Weizmann dataset, but performed poorly in UCFsports and UCF-101 datasets.

On the other hand, self-supervised discriminator achieved IS growth in all datasets, and obtains the best of all comparison models, especially in terms of FVD. We suggest that self-supervised discriminators not only help GANs learn stably in more challenging datasets ,but also show the effectiveness on both unconditional and conditional video synthesis.

At last, the categorical ArrowGAN combined with all components records the highest IS and the lowest FVD scores on all three datasets. \kb{Specifically}, categorical ArrowGAN achieves the state-of-the-art IS of 28.97 and the FVD value of 26.68 on UCF-101 which is a common benchmark for video generation. As a result, we show that our categorical ArrowGAN generates various and realistic video samples similar to real data distribution and outperforms existing advanced methods.

\subsubsection{Qualitative ablation study}
We also analyze the validity of individual components of categorical ArrowGAN qualitatively on \Fref{fig:ablation}. The most top three rows are three different videos from \textit{surfing} class in UCF-101 dataset with three different input noises and middle and bottom rows are likewise from \textit{wave}, and \textit{walk} classes in Weizmann dataset. Four frames in each row are temporally ordered from left to right.
From \Fref{fig:ablation}(a) to \Fref{fig:ablation}(d), we attach each component~\eg, conditional batch normalization \&~ projection method, diversity-sensitive regularizer and AoT discriminator on categorical MoCoGAN progressively.

As shown in the first column, categorical MoCoGAN is hard to generate recognizable videos corresponding to a given condition. Also it suffers from intra-class mode collapse that generates only a few modes of videos in a given condition. \Fref{fig:ablation}(b) describes that the projection discriminator helps to generate videos that can easily know a given class information, yet cannot generate various video samples with prominent motion.

\Fref{fig:ablation}(c) depicts that adding diversity sensitive regularizer further enforces diversity within each class. But some videos still are hard to detect objects. Lastly, \Fref{fig:ablation}(d) shows that adding AoT improves temporal coherency between frames and shape of humans, surfing and walking humans can be observed even changing their locations. Especially in surfing class, we can observe the wave moving and surfer goes along with the waves. In wave and walk classes, the change of objects' motion is more dramatic and realistic than earlier ones.

\section{Conclusion}
In this paper, we proposed ArrowGAN framework, where arrow-of-time discriminator provides extra advice for the generators to synthesize more plausible videos that runs forward in time.

Our video discriminator, Arrow-D, not only distinguishes the generated samples but also learns the temporal properties of videos without additional labeled data. Additionally, we showed that the auxiliary task should be carefully chosen regarding the target domains, \ie, learning arrow of time only improves the performance and other self-supervisory tasks do not. Furthermore, our proposed self-supervision task, AoT, shows the scalability in any pixel-level videoGANs.

Also we have succeeded in generating categorical videos based on the projection method to mitigate intra-class mode dropping. Moreover, we proposed a new baseline,~\textit{categorical ArrowGAN}, for class-conditional video generation by extending superior techniques in the image synthesis to video domain. Through experimental results, we show that our model outperforms the current state-of-the-art methods both qualitatively and quantitatively.

\section{Acknowledgement}
This work was supported by the National Research Foundation of Korea (NRF) grant funded by the Korea government (MSIT) (No.2019R1A2C2003760) and Institute for Information \& communications Technology Planning \& Evaluation (IITP) grant funded by the Korea government (MSIT) (No.2020-0-01361, Artificial Intelligence Graduate School Program(YONSEI UNIVERSITY)).

\printcredits

\bibliographystyle{cas-model2-names}

\bibliography{cas-refs}

\bio{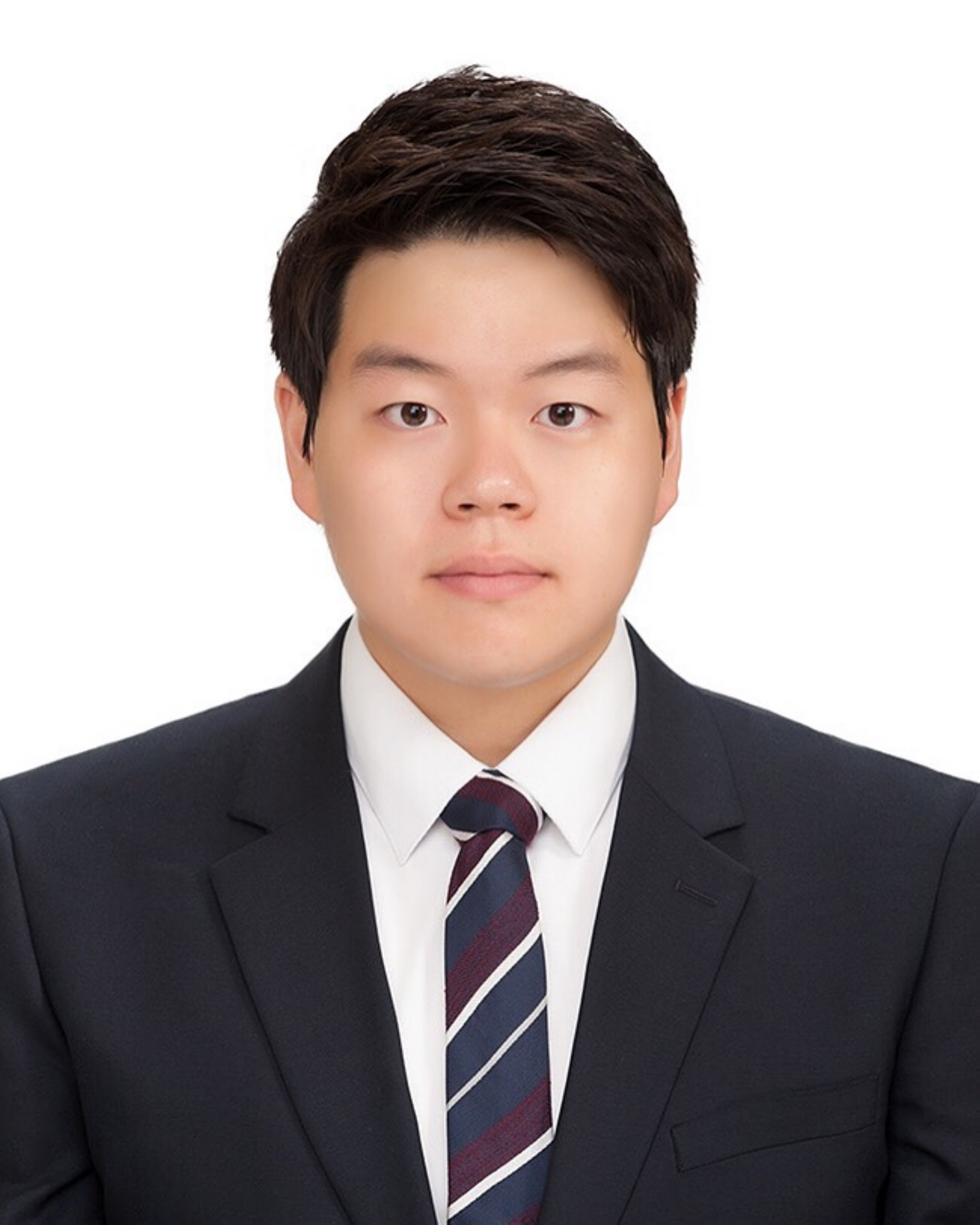}
\textbf{Kibeom Hong} is currently a Ph.D. student in Computer Science at Yonsei University, Seoul, Korea. He received the B.S. degree in Computer Science from Yonsei University, Seoul, Korea.  His research interests include generative adversarial networks, generative models, and neural style transfer..
\endbio

\bio{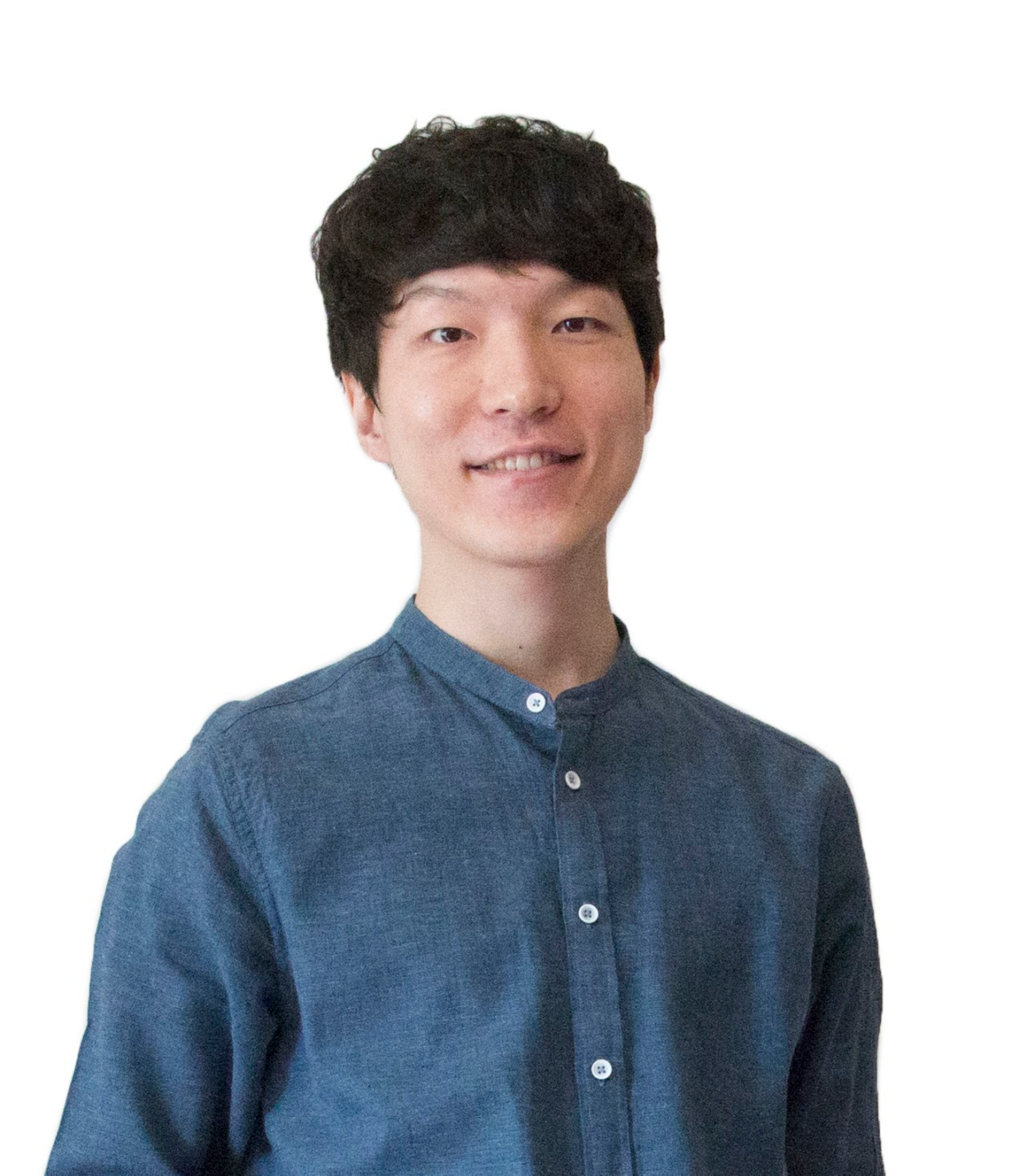}
\textbf{Youngjung Uh} is currently a researcher at Clova AI Research, Naver, Korea. He received the M.S. and the Ph.D. degrees in computer science from Yonsei University, Seoul, Korea in 2018. His research interests include generative adversarial networks, 3D reconstruction, neural style transfer and learning representation.
\endbio

\bio{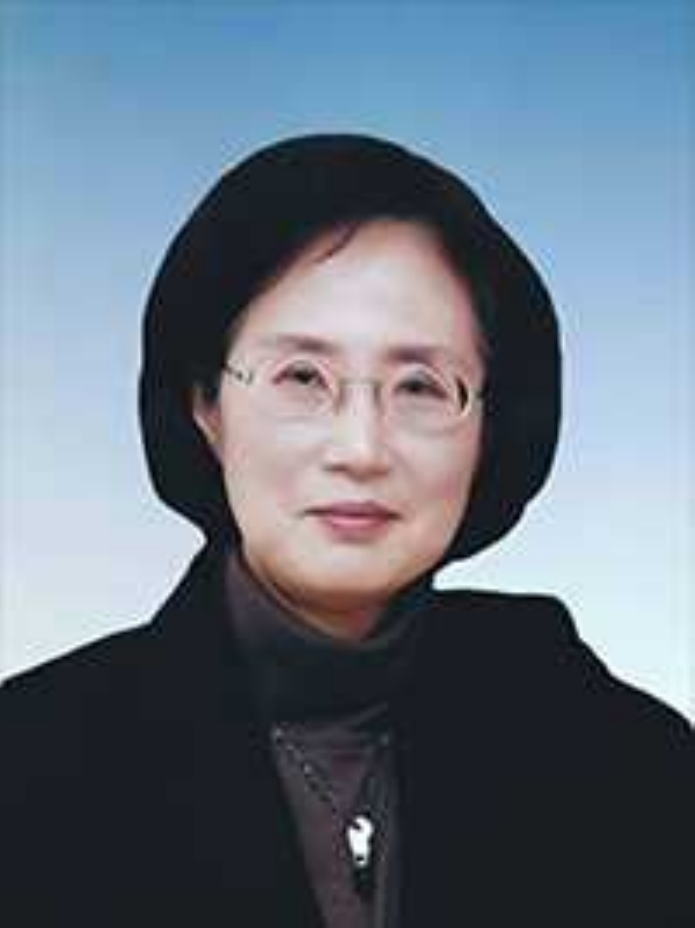}
\textbf{Hyeran Byun} is currently a professor of Computer Science at Yonsei University. She was an Assistant Professor at Hallym University, Chooncheon, Korea, from 1994 to 1995.  She served as a non executive director of National IT Indus-try Promotion Agency (NIPA) from Mar. 2014 to Feb. 2018. She is a member of National Academy Engineering of Korea. Her research interests include computer vision, image and video processing, deep learning, artificial intelligence, machine learning, and pattern recognition. She received the B.S. and M.S. degrees in mathematics from Yonsei University, Seoul, Korea, and the Ph.D. degree in computer science from Purdue University, West Lafayette, IN, USA.
\endbio

\end{document}